\documentclass[10pt,twocolumn,letterpaper]{article}

\usepackage{cvpr}              % To produce the CAMERA-READY version
\definecolor{cvprblue}{rgb}{0.21,0.49,0.74}
\usepackage[pagebackref,breaklinks,colorlinks,allcolors=cvprblue]{hyperref}

\def\confName{CVPR}
\def\confYear{2027}

\usepackage{amsmath,amsfonts,amssymb}
\usepackage{algorithmic}
\usepackage{algorithm}
\usepackage{array}
\usepackage{caption}
\usepackage{subcaption}
\usepackage{textcomp}
\usepackage{stfloats}
\usepackage{url}
\usepackage{verbatim}
\usepackage{graphicx}
\usepackage{comment}
\usepackage{bbm}
\usepackage{multirow}
\usepackage{booktabs}
\usepackage{media9}
\usepackage[table]{xcolor}

\title{MemWM: Associative Memory-Enhanced Latent World Models \\ for Long-Horizon Planning}

\author{\normalsize Yuwei Sun\textsuperscript{1}, Siyu Zhu\textsuperscript{1,2} \\
  \normalsize\textsuperscript{1}Shanghai Academy of AI for Science, 
  \textsuperscript{2}Fudan University
}

\begin{document}
\maketitle

\begin{abstract}
World models trained via latent dynamics offer a scalable path for embodied planning, yet existing methods suffer from representational collapse, computational overhead, and limited long-horizon capability. We introduce Memory World Model (MemWM), an associative memory-enhanced latent world model that augments JEPA-style prediction with a content-addressable memory bank. MemWM learns high-level historical representations directly from observations and leverages them to condition predictions via adaptive layer normalization, complementing low-level action-conditioned dynamics within a shared latent space. To prevent memory slots from collapsing into identical representations, we propose two complementary losses that encourage structural diversity across slots and balanced utilization across timesteps. On PushT, TwoRoom, Reacher, and OGBCube, MemWM attains an average success rate of 89.9\%, matching DINO-WM (89.3\%) while using roughly 4$\times$ fewer trainable parameters and no pretrained encoder, and improving over LeWM by 4.1\%. Additionally, we extend the horizon to 50 timesteps, double the 25 timesteps of the original evaluation, demonstrating MemWM's robustness to long-horizon generalization.
\end{abstract}

\section{Introduction}

World models \cite{ha2018worldmodels,nagabandi2020deep,du2023learning,micheli23tran,ko2024learning,hansen2024td,ding2024diffusion,maes2026leworldmodel} learn to predict the consequences of actions, enabling agents to plan and improve entirely within an internal model of the world. Planning with world models has become an effective framework for embodied decision-making, in which the agent imagines future trajectories given an initial observation. This demands a tight coupling of perception, prediction, and decision-making. Reconstruction-based world models learn environment dynamics by optimizing objectives that reconstruct observations from compressed latent states \cite{ha2018worldmodels,hafner2020dreamer}. However, reconstruction losses bias representations toward low-level perceptual details limiting abstraction, and increase computational cost. Moreover, relying solely on visual targets fails to exploit the richer guidance the environment provides.

Recent advances in representation and latent-space learning have further improved scalability, enabling world models to be trained directly from high-dimensional inputs. Joint-Embedding Predictive Architectures (JEPA) \cite{lecun22apath,dengsheng2025denoising,zhang2026thinkjepa,chen2025vl} compress observations into a compact, low-dimensional latent space and model temporal dynamics by predicting the latent representations of future observations. However, a central challenge remains in using JEPA for world models while preventing collapse, in which the model maps all inputs to nearly identical representations. DINO-WM \cite{zhou2024dino} employs a pretrained DINOv2 model \cite{caron2021emerging,oquab2023dinov2} to extract observation embeddings, addressing collapse at the cost of high computational overhead and reliance on external pretraining. LeWorldModel \cite{maes2026leworldmodel} tackles this problem with a lightweight latent prediction framework that uses the SigREG loss to prevent representational collapse. While LeWorldModel is efficient, it cannot match the performance of pretrained encoder-based DINO-WM.

A common misconception in model-based planning is that extending the rollout length is sufficient to elicit long-horizon behavior. In practice, simply increasing the maximum step count only causes the agent to jitter uselessly around the same goal, without understanding the underlying temporal dynamics. A recent approach layers high-level temporal abstraction and semantic structure on top of the environment's low-level contact dynamics. By reasoning at multiple temporal scales, agents plan coarsely over extended horizons while maintaining fine-grained control at shorter ones. Existing studies, such as hierarchical world models \cite{zhang2026hierarchical}, introduce a hierarchy that captures long-range dynamics while doubling the parameter count. 

An efficient strategy for such long-horizon control is to incorporate content-addressable memory \cite{turing,chan18,sun2025associa} into the decision-making structure. To overcome these limitations of conventional latent world models, we propose Memory World Model (MemWM), an associative memory-enhanced latent world model built on a content-addressable memory bank. MemWM learns reusable long-horizon contextual information from historical observations. A multi-scale shared latent space then couples these long-horizon information to low-level contact dynamics, without sacrificing model efficiency. In particular, we leverage the learned memory contents to perform long-horizon conditioning during prediction via Adaptive Layer Normalization, complementing the low-level action-conditioned prediction. Furthermore, to prevent collapse in the memory learning process, we devise two memory loss objectives that encourage both structural diversity across memory slots and balanced slot usage across timesteps.

We conduct experiments on a wide range of benchmarks including PushT, TwoRoom, Reacher, and OGBCube, along with comprehensive ablation studies on the associative memory mechanism and conditioning method. To stress-test robustness to longer horizons, we further evaluate an extended-horizon regime. Extensive empirical results demonstrate that our method matches the strongest DINO-WM baseline while using roughly 4$\times$ fewer trainable parameters and no pretrained encoder, outperforming lightweight latent world models of comparable size.

Overall, our main contributions are three-fold:

1) We propose associative memory-enhanced latent world models, MemWM, for action sequence prediction and planning. To the best of our knowledge, this is the first work to integrate a content-addressable memory bank into JEPA-style latent world models for long-horizon prediction (Section \ref{sec:method}).

2) To prevent memory slots from collapsing into identical representations, we introduce two complementary loss objectives that encourage memory slot diversification and balanced slot utilization across timesteps (Section \ref{sec:loss}).

3) Extensive experiments across a broad range of tasks demonstrate that our method outperforms conventional latent world models including PLDM, DINO-WM, and LeWorldModel in aggregate success rate. We further analyze memory slot usage to characterize the functional roles of various memory slots, providing insight into the interpretability of memory contents (Section \ref{sec:exp}).

\section{Related work}

This section provides a summary of the most relevant work on world models including the reconstruction-based methods and the latent world model methods. We investigate the relatedness and limitations of these studies comparing with our memory-enhanced method.

\paragraph{Reconstruction-based world models}

Reconstruction-based world models learn latent environment dynamics by optimizing objectives that reconstruct observations from compressed latent states, encouraging representations to retain sufficient information for predicting future outcomes. Early approaches such as World Models \cite{ha2018worldmodels,micheli23tran} combined variational latent representations with recurrent dynamics and pixel reconstruction to enable imagined rollouts for control. Subsequent methods, including PlaNet \cite{hafner2019planet}, Dreamer \cite{hafner2020dreamer}, DreamerV2 \cite{hafner2021dreamerv2}, and DreamerV3 \cite{hafner2023dreamerv3}, leveraged recurrent state-space models trained with observation reconstruction and latent transition objectives to learn compact world models. While these methods demonstrated competitive empirical performance, reconstruction losses bias representations toward low-level perceptual details and greatly increase computational cost.

\paragraph{Latent world models}

Joint-Embedding Predictive Architecture (JEPA) \cite{lecun22apath,bardes2024v,assran2023self,dong2024brain,dengsheng2025denoising,assran2025v,munim2026echojepa,zhang2026thinkjepa,chen2025vl} is a self-supervised learning method for world modeling that learns predictive representations in latent space rather than reconstructing low-level pixels. For instance, DINO-WM \cite{zhou2024dino} leverages spatial patch features pre-trained with DINOv2 \cite{caron2021emerging,oquab2023dinov2} in latent space, enabling more precise planning and control. PLDM \cite{sobal2022joint,sobal2025stre,sobal2025learn} learns representations using VICReg \cite{bardes2022vicreg} at the cost of training instabilities and scalability limitations. Hierarchical World Models \cite{zhang2026hierarchical} provide a learning framework with separate short and long-horizon modules that enhances the performance while doubling the parameter count. LeWorldModel \cite{maes2026leworldmodel} is a compact JEPA-style world model that predicts next-frame embeddings, however its performance degrades over long horizons and does not match DINO-WM with the pretrained DINOv2 encoder. In contrast, our MemWM leverages a memory bank for learning long-horizon contexts, improving performance without additional prediction modules or a pretrained encoder.

\section{Method}

\begin{figure*}
    \centering
    \includegraphics[width=0.95\linewidth]{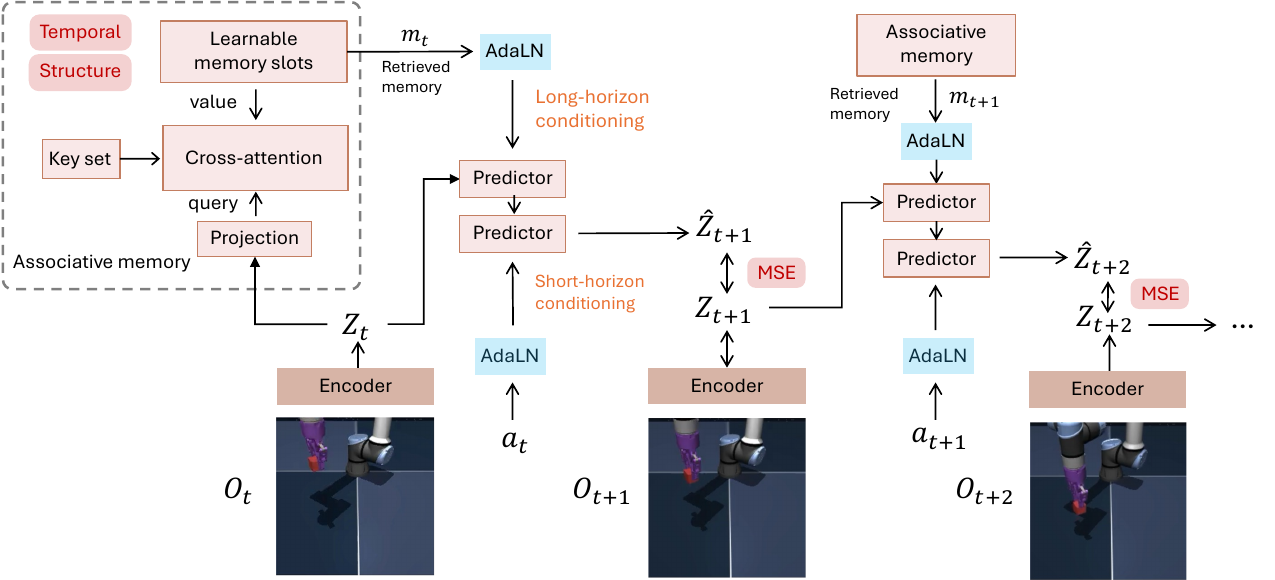}
    \caption{Overview of MemWM. The encoder maps an observation to a latent $z_t$. Then, the memory bank retrieves a context $m_t$ via content-addressable attention. Finally, the predictor employs sequential AdaLN conditioning to produce the next time step latent $\hat{z}_{t+1}$.}
    \label{fig:scheme}
\end{figure*}

In this section, we present the associative memory-enhanced latent world model architecture MemWM (Figure \ref{fig:scheme}). We begin by detailing the joint embedding predictive architecture (JEPA), then introduce a new content-addressable memory component that learns diverse memory slot representations for long-horizon conditioning. The predictor then leverages both the low-level contact information of actions and the long-horizon contextual information retrieved from the memory bank to forecast the next latent state. Furthermore, we propose two types of memory loss objectives that encourage slot diversity from both structural and temporal perspectives, preventing memory learning collapse. We employ model predictive control during inference to enable effective action sequence rollouts.

\subsection{JEPA-style latent world models}

A world model is a learned dynamical model that approximates how an environment $o_t$ evolves in response to an agent’s actions $a_t$, estimating transition dynamics of the form $p(o_{t+1} | o_t, a_t)$. This work builds upon the joint embedding predictive architecture (JEPA) \cite{dengsheng2025denoising,zhang2026thinkjepa,maes2026leworldmodel}, where an encoder $\text{enc}_\theta(\cdot)$ such as a Vision Transformer (ViT) \cite{dosovitskiy2020image} maps a given frame $o_t$ into a low-dimensional latent embedding $z_t \in \mathbb{R}^D:\text{enc}_\theta(o_t)$. A predictor aims to model the environment dynamics in latent space $p(z_{t+1} | z_t, a_t)$ by predicting the latent $\hat{z}_{t+1}$ of the next frame given the latent $z_t$ and an action $a_t$. The action is incorporated into the predictor as a condition through Adaptive Layer Normalization (AdaLN) as follows: 
\begin{equation}
\boldsymbol{\gamma}_s, \boldsymbol{\beta}_s, \boldsymbol{\alpha}_s, \boldsymbol{\gamma}_f, \boldsymbol{\beta}_f, \boldsymbol{\alpha}_f = \Phi(a_t),
\end{equation}
where $\Phi$ is a linear projection and $\boldsymbol{\gamma}_s, \boldsymbol{\beta}_s, \boldsymbol{\alpha}_s, \boldsymbol{\gamma}_f, \boldsymbol{\beta}_f, \boldsymbol{\alpha}_f$ are adaptive parameters. The conditioned latent output $z_{t+1}$ is defined as:
\begin{equation}
\text{AdaLN}(z_t, \boldsymbol{\gamma}, \boldsymbol{\beta}) = \boldsymbol{\gamma} \odot \text{LN}(z_t) + \boldsymbol{\beta},
\end{equation}
\begin{equation}
z_t' = z_t + \boldsymbol{\alpha}_s \odot \text{Attn}\!\left(\text{AdaLN}(z_t, \boldsymbol{\gamma}_s, \boldsymbol{\beta}_s)\right), 
\end{equation}
\begin{equation}
\hat{z}_{t+1} = z_t' + \boldsymbol{\alpha}_f \odot \text{MLP}\!\left(\text{AdaLN}(z_t', \boldsymbol{\gamma}_f, \boldsymbol{\beta}_f)\right),
\end{equation}
where $\text{LN}$ is LayerNorm, Attn is the attention layer, MLP is the multilayer perceptron, and $\odot$ is the element-wise multiplication.

\subsection{Associative memory for long-horizon world modeling}
\label{sec:method}

Conventional latent world models including LeWorldModel struggle with long-horizon prediction tasks. One reason is that they learn latent dynamics solely through action-conditioned transitions. Training only on short-horizon contact information undermines the model's planning performance. To overcome this limitation, an intuitive idea is to induce long-term memory during prediction. MemWM leverages a learnable associative memory mechanism for long-horizon world modeling in JEPA-style latent world models, by accumulating historical latents to provide diversified contextual information. This high-level contextual information is then fed into the predictor along with the action to enhance long-horizon prediction.

\subsubsection{Associative memory structure}

Humans constantly rely on associative memory to predict the surrounding world. The content-addressable capability of associative memory has been employed to enhance long-range input association in modern transformer architectures \cite{sun2025associa}. Leveraging such learnable memory disentangles information across timesteps and inputs, enabling latent predictions that combine short-horizon actions with long-horizon memory contexts to enhance agent planning. 

In particular, an associative memory bank is comprised of $M$ learnable low-rank slots $\Gamma \in \mathbb{R}^{M \times D}$. Each memory slot diversifies into a different general assumption about observations and environment dynamics, corresponding to various planning stages such as moving toward the object, pushing the object, and moving to a specific area. We elaborate on memory updating, content retrieval, and long-horizon conditioning in the remainder of this section.

\subsubsection{Learnable memory bank and content retrieval}

The learnable memory slots $\Gamma$ are paired with a set of learnable keys $\Xi  \in \mathbb{R}^{M\times D}$ that serve as indexes for writing to the memory bank, with each key and memory slot forming a pair. We decouple the training of keys from that of memory slots, ensuring that learning the keys does not substantially interfere with the memory update. These learnable keys are utilized to compute cross-attention with the observation latents, which assigns different weights to each memory slot accordingly. The attention weights for input latent states $z_t$ can be computed as follows:
\begin{equation}
A(z_t, \Xi) = \text{softmax}\left(\frac{(z_t W^Q) \Xi^\top}{\sqrt{D}\cdot \tau}\right),
\end{equation}
where $W^Q \in \mathbb{R}^{D \times D}$ is a query projection matrix and $\tau$ is the retrieval temperature. We discuss the selection of the retrieval temperature in Section \ref{sec:mem}.

The attention weights represent the similarity between a specific observation latent state and all keys. Depending on these weights, the memory slots corresponding to the specific keys are leveraged to compute the final weighted memory retrieval result $m_t = \sum_{i=1}^M A_i\,\Gamma_i$. The retrieved memory $m_t$ is then employed as long-horizon conditioning to guide predictor training. %Through end-to-end training, we aim to learn a set of diverse memory slots that represent different scenarios, such as approaching the object, pushing the object, and moving to the target area. The keys are also learned during training, as the indices to identify memory slots.

\subsubsection{Long-horizon prediction conditioning}

To introduce the additional condition of memory information into the world model, we leverage another Adaptive Layer Normalization (AdaLN) module. There are several choices for the conditioning design. For instance, we can adopt either a parallel or a sequential design, and the AdaLN modules can either correspond to two separate predictors for the short-horizon and long-horizon conditions, respectively, or correspond to the same predictor using parameter sharing. Comparing various combinations, we found that employing a sequential design with the memory condition coming first, while using parameter sharing, contributes to the best prediction accuracy and the reduced parameter size (Table \ref{tab:condition}). Consequently, the predictor $\mathrm{Pred}(\cdot)$ first observes the long-horizon condition of the retrieved memory $m_t$, outputting an intermediate latent $z_t^\text{int}$, and then leverages the short-horizon condition of actions $a_t$, yielding the final output $\hat{z}_{t+1}$ as follows:
\begin{equation}
z_t^\text{int} = \text{Pred}(z_t, m_t),
\end{equation}
\begin{equation}
\hat{z}_{t+1} = \text{Pred}(z_t^\text{int}, a_t).
\end{equation}

\subsection{Training objectives}

Training objectives include a prediction loss for latent error minimization and two types of memory losses that enforce structural and temporal diversity on the memory slot learning, respectively.

\subsubsection{Prediction loss}

The prediction loss aims to minimize the reconstruction error at each future step. In particular, given a latent representation $z_t$ and action $a_t$, the model predicts the next latent state $\hat{z}_{t+1} = \text{Pred}(z_t|a_t)$. The prediction loss over a horizon of $T$ future steps is defined as:
\begin{equation}
\mathcal{L}_{\text{mse}} = \frac{1}{T} \sum_{k=1}^{T} \left\| \hat{z}_{t+k} - z_{t+k} \right\|_2^2.
\end{equation}
Where $\mathcal{L}_{\text{mse}} $ is the mean squared error loss. Furthermore, to prevent representation collapse and maintain a well-conditioned latent space, we leverage the SIGReg loss \cite{maes2026leworldmodel} at each step over the predicted latents $\hat{Z}$, i.e., $\text{SIGReg}(\hat{Z})$. The prediction loss can be defined as $\mathcal{L}_{\text{pred}} = \mathcal{L}_{\text{mse}} + \lambda_{\text{sigreg}} \, \text{SIGReg}(\hat{Z})$, where $\lambda_{\text{sigreg}}$ is a weighting coefficient controlling the strength of the SIGReg regularization.

\subsubsection{Associative memory loss}
\label{sec:loss}

We propose two complementary memory losses to prevent the collapse of memory learning, encouraging diversity from both structural and temporal perspectives: (1) a slot diversity loss measuring structural differences among memory slots, and (2) a temporal diversity loss measuring temporal differences in retrieval results across timesteps.

\paragraph{Slot diversity loss}

The slot diversity loss aims to diversify associative memory slots preventing structural collapse. It enforces smoothness over slot distributions by penalizing pairwise similarity, preventing memory slots from degenerating into identical information and ensuring each slot captures unique facets of the data distribution. In particular, the slot diversity loss is defined as follows:
\begin{equation}
\mathcal{L}_{\text{div}} = \text{SIGReg}(\Gamma) 
= \sum_{i \neq j} \left( \frac{\Gamma_i^\top \Gamma_j}{\|\Gamma_i\| \|\Gamma_j\|} \right)^2,
\end{equation}
where $\Gamma_i$ denotes the \textit{i}-th memory slot.

\paragraph{Temporal diversity loss}

The temporal diversity loss $\mathcal{L}_{\text{tem}}$ measures temporal differences in memory retrieval results across time steps by computing the entropy of slots' attention distribution over a horizon of $T$ future steps. In particular, the average attention weights assigned to a specific slot $i$ across a horizon of $T$ is $\bar{A}_i = \frac{1}{T} \sum_{k=1}^{T} A_{i}^{(k)}$. Then, the entropy of associative memory is defined as:
\begin{equation}
H_{\text{slot}} = -\sum_{i=1}^{M} \bar{A}_i \log(\bar{A}_i + \epsilon),
\end{equation}
where $\epsilon = 10^{-8}$ is a small constant for numerical stability. The entropy is maximized when the memory slots are used uniformly across the $T$ future steps, i.e., $H_{\text{max}} = \log M$. We formulate the normalized temporal diversity loss as:
\begin{equation}
\mathcal{L}_{\text{tem}} = 1 - \frac{H_{\text{slot}}}{H_{\text{max}}} = 1 + \frac{1}{\log M} \sum_{i=1}^{M} \bar{A}_i \log(\bar{A}_i + \epsilon),
\end{equation}
which approaches zero when the entropy reaches its theoretical maximum $\log M$. All the memory slots are leveraged more uniformly across time encouraging temporal diversity. 

Finally, the complete memory loss combines both objectives: $\mathcal{L}_{\text{memory}} = \mathcal{L}_{\text{div}} + \lambda_{\text{tem}} \, \mathcal{L}_{\text{tem}},$ where $\lambda_{\text{tem}}$ is a balancing coefficient that controls the relative importance of the temporal diversity constraint. The overall training objective consists of the prediction loss and the associative memory loss: $\mathcal{L}_{\text{total}} = \mathcal{L}_{\text{pred}} + \mathcal{L}_{\text{memory}}$. Expanding this gives the complete formulation:
\begin{equation}
\begin{aligned}
\mathcal{L}_{\text{total}} &= 
\underbrace{\mathcal{L}_{\text{mse}} + \lambda_{\text{sigreg}} \, \text{SIGReg}(\hat{Z})}_{\text{Prediction Objective}} \\
&+ \underbrace{\text{SIGReg}(\Gamma) + \lambda_{\text{tem}} \, (1 - \frac{H_{\text{slot}}}{H_{\text{max}}})}_{\text{Associative Memory Objective}}.
\end{aligned}
\end{equation}

\subsection{Inference-time planning}

%During inference, we leverage the Model Predictive Control (MPC) \cite{zhou2024dino,sobal2025learn,maes2026leworldmodel} to predict and optimize the control action sequence over a finite horizon $T$. In particular, planning starts from the current latent state \( z_t \), samples a set of candidate actions from a Gaussian distribution $\delta: \mathcal{N}\!\left(\mu^{(t)}, \sigma^{(t)}\right)$, and rolls them through the memory world model for sequences of future latent states $\{\hat{z}_k\}_{k=t+1}^{T}$. We evaluate each sequence with a cost function relative to the target state based on the cross-entropy loss, select the top-performing trajectories and update the action selection distribution by recomputing its mean $\mu$ and covariance $\sigma$. For the stable update, we select only the first actions \( \{a_n^{*(t)}\}_{n=1}^N\) in the sequences at each step: 
%\begin{align}
%\mu^{(t+1)} &= \frac{1}{N}\sum_{n=1}^{N} a_n^{*(t)}, \\
%\sigma^{(t+1)} &= \frac{1}{N}\sum_{n=1}^{N}\left(a_n^{*(t)} - \mu^{(t+1)}\right)^2,
%\end{align}
%where $N$ is the number of total samples drawn from the Gaussian distribution. This process thereby concentrates future samples around better action candidates until an optimal action sequence is obtained.

During inference, we leverage Model Predictive Control (MPC) \cite{hansen2022temporal,zhou2024dino,sobal2025learn,maes2026leworldmodel} to predict and optimize the control action sequence over a finite horizon $T$. The goal image is encoded with the same encoder employed during training to obtain a goal latent $z_g=\text{enc}_\theta(o_g)$. In particular, planning starts from the current latent state $\hat{z}_1=\text{enc}_\theta(o_1)$, samples $N$ sets of candidate actions $\{a_{1:T}^n\}_{n=1}^N$ randomly from a Gaussian distribution $\mathcal{N}\!\left(\mu^{(t)}, \sigma^{(t)}\right)$, and rolls them out through the memory world model to obtain sequences of future latent states $\{\hat{z}_k\}_{k=t+1}^{T}$. The model predicts latent transitions according to $\hat{z}_{t+1} = \text{Pred}(\hat{z}_t, a_t)$.

We evaluate each sequence with a cost function measuring the distance between the predicted final latent $\hat{z}_T$ and the goal latent $z_g$, based on the squared $\ell_2$ loss $\mathcal{C}(\hat{z}_T)=\|\hat{z}_T-z_g\|_2^2$. We then select the $U$ top-performing trajectories $\{a_{1:T}^{*(n)}\}_{n=1}^U$, i.e., those minimizing $\mathcal{C}(\hat{z}_T)$, and update the action selection distribution by recomputing its mean $\mu$ and covariance $\sigma$. For a stable update, we select only the first action of each sequence $\{a_1^{*(n)}\}_{n=1}^U$ at each step:
\begin{align}
\mu^{(t+1)} &= \frac{1}{U}\sum_{n=1}^{U} a_1^{*(n)}, \\
\sigma^{(t+1)} &= \frac{1}{U}\sum_{n=1}^{U}\left(a_1^{*(n)} - \mu^{(t+1)}\right)^2.
\end{align}
This process concentrates future samples around better action candidates until an optimal action sequence is obtained.

\section{Experiments}
\label{sec:exp}

In this section, we present the settings and extensive experimental results on various 2D and 3D planning tasks. We compare Memory World Model (MemWM) with a broad range of conventional latent world models, including comprehensive ablation studies on the effectiveness of the associative memory mechanism, the memory diversity losses, and the conditioning mechanism. We further provide detailed visualizations of planning trajectories and memory slot usage distributions, demonstrating the interpretability of memory slot roles in various stages of the planning tasks. %In addition, we provide a tailored dataset to evaluate the long-horizon prediction performance of latent world models.

\paragraph{Datasets}

We evaluate on a diverse set of tasks including navigation, motion planning and manipulation. All datasets are from the stable-worldmodel \cite{maes2026stable} framework. In particular, PushT is a robotic manipulation task where the goal is to push a T-shaped block to target positions using top-down visual observations. TwoRoom is a 2D navigation task where an agent moves through two connected rooms to reach a specified location. Reacher is a continuous control task involving a robotic arm that moves its end-effector to randomly appearing target points. OGBCube is a 3D manipulation task requiring an agent to perform tasks like pushing or picking a cube. We provide more details in Appendix \ref{app:dataset}.

\paragraph{Settings}

For the reimplementation of all baseline methods, we employed the author-recommended settings and hyperparameters. For the proposed MemWM model, we employed a batch size of 128, a memory slot number of 10, and a dimension of 192 for each memory slot. We initialized the learnable memory slots and keys with random values scaled by 0.02 and employed a query projector with two dense layers and an intermediate dimension of 192. The encoder is a vision transformer \cite{dosovitskiy2020image} and the observation embedding $z_t$ is constructed from the [CLS] token from the last layer of the vision transformer. The predictor is a 6-layer transformer with 16 attention heads. Please refer to Appendix \ref{app:hype} for the complete model settings and hyperparameters.

\paragraph{Baselines}

We compare our method to the following conventional latent world models:

\begin{itemize}
    \item PLDM \cite{sobal2022joint,sobal2025learn}: learns representations end-to-end with additional regularization terms, at the cost of training instability and scalability limitations.
    \item DINO-WM \cite{zhou2024dino}: employs pretrained DINOv2 \cite{caron2021emerging,oquab2023dinov2} as the encoder and additional proprioceptive inputs to mitigate representation collapse in latent space training.
    %\item Hierarchical World Model (HWM) \cite{zhang2026hierarchical}: employs high-level conditioning via additional predictors to improve long-horizon planning, though this doubles the parameter count.
    \item LeWM \cite{maes2026leworldmodel}: is a compact JEPA-style world model that predicts next-frame embeddings. Ablating the associative memory mechanism from MemWM recovers the LeWM architecture.
\end{itemize}

\paragraph{Latent planning performance}

Given an initial observation and a goal, the agent model aims to output a sequence of actions to reach the goal. The predictor rolls out future latents with a selected action sequence updated with model predictive control, where a cost between the final predicted latent and the goal embedding guides a solver to optimize actions. Moreover, we assume the goal is reachable within 25 steps and employ a maximum planning budget of 50 steps for all tasks. We report the mean success rate across three random seeds, evaluated on the same set of 50 trajectories (Table \ref{tab:baseline}). 
%MemWM achieves the highest average success rate (89.9\%), surpassing DINO-WM (89.3\%), LeWM (85.8\%), and PLDM (79.5\%). DINO-WM performs best in Two-Room (100\%) and OGBCube (86\%). While no single baseline dominates all tasks, MemWM exhibits the most consistent and well-rounded performance. 
DINO-WM remains strongest on Two-Room (100\%) and OGBCube (86\%), where its large pretrained visual backbone provides a clear advantage for fine-grained spatial reasoning. MemWM is the best on PushT (96.7\%) and Reacher (90.7\%). While no single baseline dominates all tasks, MemWM attains the highest average success rate (89.9\%) comparable to DINO-WM (89.3\%), while using 3.95$\times$ fewer trainable parameters (19M vs.\ 75M) and without relying on a frozen 1.1B-parameter DINOv2 encoder. Relative to LeWM (18M params, 85.8\%), MemWM improves the average by 4.1\%. 
%Additionally, LeWM employed a larger evaluation budget of 100 on the TwoRoom task. Under the same 50-step budget, its performance would degrade further to 40\%.

%Our method not only achieves superior performance over the various baselines but also uses fewer parameters than PLDM and DINO-WM (Table \ref{tab:baseline}), owing to its efficient memory-based conditioning mechanism.MemWM attains the highest average success rate (89.9\%) and is comparable to DINO-WM (89.3\%), while using roughly 4$\times$ fewer trainable parameters (19M vs. 75M) and without relying on a frozen 1.1B-parameter DINOv2 encoder. MemWM matches DINO-WM's aggregate performance with a lightweight, end-to-end trained encoder and a compact memory-based conditioning mechanism. Furthermore, LeWM utilized a large evaluation budget in the TwoRoom task \cite{maes2026leworldmodel}. Using the same 50-step budget as the other methods generally degrades its performance. In contrast, our method yields better performance in the TwoRoom task even with the limited planning budget. In addition, compared to LeWM with 18M parameters, MemWM has 19M parameters while improving performance by 4.1\%, outperforming DINO-WM built on the pretrained DINOv2 encoder.

\begin{table*}[!t]
    \small
    \centering
    \renewcommand{\arraystretch}{1.12}
        \caption{Performance comparison on planning tasks (Success Rate \(\uparrow\)). Parameter counts report trainable parameters only. DINO-WM relies on a frozen 1.1B-parameter DINOv2 encoder not included in its count. Best per column in \textbf{bold} and second-best \underline{underlined}.}
    \begin{tabular}{l|cccccc}
        \toprule
        \rowcolor{CornflowerBlue!10}
        Model & {Push-T (\%)} & {Two-Room (\%)} & {Reacher (\%)} & {OGBCube (\%)} & {Average (\%)} & {Params (M)}\\
        \hline
        PLDM \cite{sobal2022joint,sobal2025learn}  & 78.0 $\pm$ 5.0  & \underline{97.0 $\pm$ 0.94} & 78.0 $\pm$ 5.85 & 65.0 $\pm$ 2.49 & 79.5 & 20\\
        LeWM \cite{maes2026leworldmodel} & \underline{96.0 $\pm$ 2.83} & 87.0 $\pm$ 1.79 & \underline{86.0 $\pm$ 5.25} & 74.0 $\pm$ 2.49 & 85.8 & 18\\
        %HierarchicalWM \cite{zhang2026hierarchical} & 89 &&&&& 100\\
        DINO-WM \cite{zhou2024dino}  & 92.0 $\pm$ 1.63 & \textbf{100} & 79.0 $\pm$ 5.85 & \textbf{86.0 $\pm$ 2.49} & \underline{89.3} & 75 (+1.1B DINOv2 encoder)\\
        MemWM (Ours) & \textbf{96.7 $\pm$ 0.94} & 90.7 $\pm$ 1.89 & \textbf{90.7 $\pm$ 0.94} & \underline{81.3 $\pm$ 1.89} & \textbf{89.9} & 19 \\ %tworoom 80 epoch
        \bottomrule
    \end{tabular}
    \label{tab:baseline}
\end{table*}

\begin{figure*}
    \centering
    \includegraphics[width=0.72\linewidth]{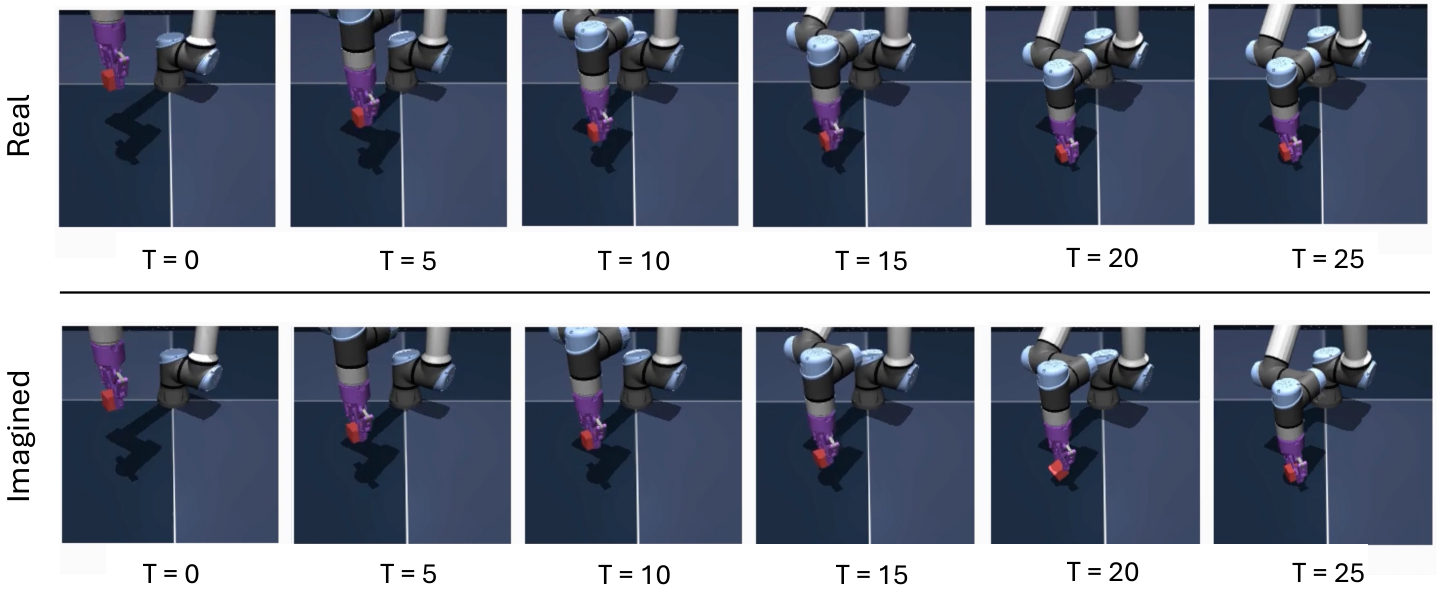}
    \caption{Trajectory visualization of MemWM latent planning in the OGBCube dataset.}
    \label{fig:rollout}
\end{figure*}

\paragraph{Trajectory visualization}

To interpret latents at each step, we train a decoder to project the compressed latent representations back into the original observation space. As shown in Figure \ref{fig:rollout}, MemWM produces stable future rollouts closely aligned with the real environment predictions.

%\begin{figure*}
%    \centering
%    \includegraphics[width=0.7\linewidth]{fig/tworoom.pdf}
%    \caption{Trajectory visualization of latent world model planning with MemWM and LeWM. MemWM demonstrates better aligned latent trajectories compared to LeWM, which closely resemble the ground truth trajectories.}
%    \label{fig:traj}
%\end{figure*}

\paragraph{Interpretability of different memory slot roles}

\begin{figure*}
    \centering
    \includegraphics[width=0.99\linewidth]{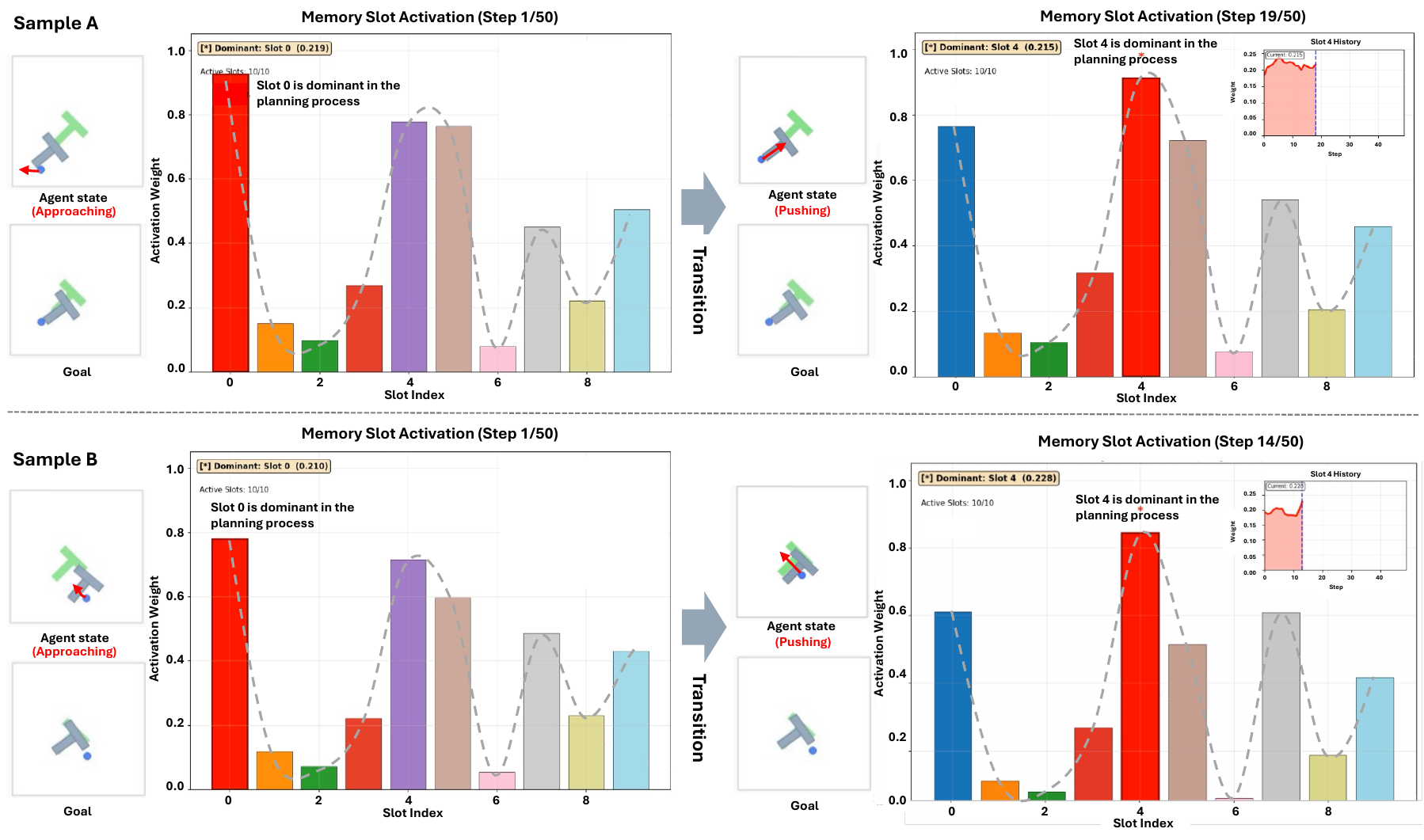}
    \vspace{-8pt}
    \caption{Visualization of memory slot usage. We observe distinct activation patterns across planning stages. In the Push-T environment, the first slot exhibits peak activation when the agent starts to approach the T-shaped block, whereas the fifth slot dominates when the agent initiates pushing the block toward the target area. We observe this pattern consistently across Push-T samples and analogous stage-specific activation patterns appear in the other tasks.}
    %This phenomenon generalizes across various samples, exhibiting similar transitions during planning.
    \label{fig:slot_usage}
\end{figure*}

The diversity of memory slots examines how memory information is leveraged throughout the planning tasks. By analyzing slot activation distributions, we aim to characterize the distinct functionalities that various slots might serve. In particular, the slot usage rates $u_k = \frac{1}{N}\sum_{i=1}^{N} \sum_{t=1}^{T} A_{k}^{(i,t)}$ are monitored across all $N$ queries, and the slot diversity is measured as $\frac{1}{\sum_{k=1}^{K} u_k^2}$, with higher values indicating more uniform usage, while active slots are defined as those with usage rate exceeding $0.01$, i.e., $N_{\text{active}} = \sum_{k=1}^{K}\mathbbm{1}[u_k > 0.01]$. We present the analysis in Figure \ref{fig:slot_usage}, where memory slots form various roles and show different strengths of activations according to the various stages of the agent. Some slots appear to be more active at an early stage when the agent is moving towards the object, while others are activated heavily when the object is being pushed. %The transition of agent states constitutes subgoals, and the memory usage pattern changes when a subgoal is reached, showing the memory bank learns distinct long-horizon knowledge of the environment dynamics. %Figure \ref{fig:slot_usage} demonstrates the subgoals inferred by the memory usage distribution, where the corresponding latents of each subgoal state are decoded for visualization and we use the dotted red columns to indicate them.

%by activating each specific memory slot, the agent predicts latents with specific motion directions. We visualize the frames showing a trasition state of the agent. The transition state presents its relation to the actions where a set of action-conditioned states are overlapped in the recontructed frame, showing the interpretability of different stages using the long-horizon conditioning method.

\paragraph{Ablation studies}

A comprehensive ablation study provides insights into the functionality of each key component in MemWM. We conduct the following ablations:
\begin{itemize}
    \item W/O shared parameters: The predictors for the memory and action conditions are separate. The observation embeddings first pass through a predictor conditioned on the retrieved memory latents, and then through another predictor conditioned on the actions. This differs from the default setting, where the predictors that process the observation embeddings conditioning on the memory and actions share parameters.
    \item W/O temporal diversity loss: The temporal diversity loss term is omitted during training.
    \item W/O memory loss: The two types of associative memory losses are removed, i.e., the regularization on both memory structural and temporal diversity is discarded.
    %\item W/O associative memory: The intact associative memory mechanism is removed, which degenerates the architecture to the LeWM architecture.
\end{itemize}

As shown in Table \ref{tab:push_t_results}, removing the temporal diversity loss leads to a non-trivial drop from 89.9\% to 86.7\%, indicating its important role in capturing temporal dynamics. Disabling both memory losses (W/O memory loss) causes a more substantial decline to 83.5\%, particularly on Reacher (90.7\% vs.~84.0\%) and OGBCube (81.3\% vs.~66.0\%), highlighting the importance of memory regularization in complex tasks. The most significant degradation occurs when the parameter sharing strategy is removed (W/O shared params), yielding an average of 82.9\%. This suggests that sharing information between short-horizon and long-horizon predictions is beneficial. Overall, these results demonstrate that each designed component contributes meaningfully to the model's prediction performance.

%The absence of parameter sharing leads to degraded performance, possibly due to confusion between long-term and short-term signals. Separating action conditioning from memory conditioning stabilizes training while enabling more accurate planning.

\begin{table*}[!t]
\small
\setlength{\tabcolsep}{6pt}
\renewcommand{\arraystretch}{1.12}
    \centering
    \caption{Performance comparison on the various ablations (Success Rate \(\uparrow\)), demonstrating the effectiveness of the proposed memory mechanism and memory loss objectives. Best per column in \textbf{bold} and second-best \underline{underlined}.}
    \begin{tabular}{l|ccccc}
        \toprule
        \rowcolor{CornflowerBlue!10}
         Model & {Push-T (\%)}& {Two-Room (\%)} & {Reacher (\%)} & {OGBCube (\%)} & Average (\%)\\
        \hline
        MemWM & \textbf{96.7 $\pm$ 0.94} & \textbf{90.7 $\pm$ 1.89} & \textbf{90.7 $\pm$ 0.94} & \textbf{81.3 $\pm$ 1.89} & \textbf{89.9}\\  
        % 98/96/96 & 92/92/88 & 90/92/90 & 84/ & 91.0 some randomness in the prediction performance cube_temporal_seed3073_
        W/O temporal diversity loss & \underline{94.7 $\pm$ 2.49} & \underline{90.0 $\pm$ 1.63} & \underline{88.7 $\pm$ 1.89} & \underline{73.3 $\pm$ 2.49} & \underline{86.7} \\ 
        % 98/92/94 & 92/88/90 & 90/90/86 & 74/70/76 & 88.0 
        % pusht_seed3072_30 pusht_seed3074_60   
        % tworoom_seed3073_90  tworoom_seed3074_90  
        % cube_seed3074_40
        % reacher_seed3074_100 reacher_seed3073_80 reacher_seed3072_100 
        W/O memory loss   & 94.7 $\pm$ 0.94 & 89.3 $\pm$ 1.89 & 84.0 $\pm$ 3.27 & 66.0 $\pm$ 3.27 & 83.5 \\%96/94 & 92/88 & 80/88/84 & 70/66 & 84.5
        W/O shared params & 90.7 $\pm$ 2.31  & 87.3 $\pm$ 3.06 & 84.7 $\pm$ 1.15 & 68.7 $\pm$ 2.31 & 82.9\\ %88/92/92  & 88/90/84 & 86/84/84 & 66/70/70
        %W/O associative memory  & 96.0 $\pm$ 2.83 & 87.0 $\pm$ 1.79 & 86.0 $\pm$ 5.25 & 74.0 $\pm$ 2.49 & 85.8\\
        %MemWM W/O memory loss and shared params & 92 & 88 & \\
        %MemWM w/ entropy memory loss &     \\
        %MemWM w/ sigreg memory loss &     \\
        \bottomrule
    \end{tabular}
    \label{tab:push_t_results}
\end{table*}

\paragraph{Discussion on memory hyperparameters}
\label{sec:mem}

We vary the number of memory slots from 5 to 20 in Table \ref{tab:mem}. Increasing the slot number and thus memory capacity does not necessarily yield better performance. Through ablation studies, we select the slot number to suit different task complexity, finding that 10 slots provide the optimal overall performance. We study the retrieval temperature $\tau$, which governs the sharpness of attention weights over the key set: smaller values sharpen the distribution toward the most relevant key, while larger values flatten it toward uniform weighting. Table \ref{tab:mem} shows that $\tau = 0.5$ yields optimal performance, preventing inactive memory slots and facilitating learning.

\begin{table*}[!t]
  \centering
  \small
  \renewcommand{\arraystretch}{1.12}
  \caption{Model performance with varying memory hyperparameters (Success Rate \(\uparrow\)). We employ a memory slot number of 10 and a retrieval temperature of 0.5 by default, which results in the optimal performance. Best per column in \textbf{bold} and second-best \underline{underlined}.}
  \begin{tabular}{lcccccc}
    \toprule
    \rowcolor{CornflowerBlue!10}
    {Hyperparameter} & & {Push-T (\%)}& {Two-Room (\%)} & {Reacher (\%)} & {OGBCube (\%)} & Average (\%)\\
    \hline
    \multirow{4}{*}{Number of Memory Slots} & \,\,\,\,\,\,5\,\,\,\,\,\, & 93.3 $\pm$ 0.94 & \underline{89.3 $\pm$ 0.94} & \underline{86.7 $\pm$ 1.89} & 76.7 $\pm$ 3.40 & \underline{86.5} \\%94/92 & 90/88 & 88/84 & 80/72/78 & 88.0\\
                                     & 10 & \textbf{96.7 $\pm$ 0.94} & \textbf{90.7 $\pm$ 1.89} & \textbf{90.7 $\pm$ 0.94} & \textbf{81.3 $\pm$ 1.89} & \textbf{89.9}\\
                                     & 15 & 91.3 $\pm$ 0.94 & \underline{89.3 $\pm$ 0.94} & 86.0 $\pm$ 1.63 & \underline{77.3 $\pm$ 0.94} & 86.0\\ 
                                     & 20 & \underline{94.0 $\pm$ 1.63} & 88.7 $\pm$ 0.94 & 85.3 $\pm$ 2.49 & 74.7 $\pm$ 0.94 & 85.7\\ %3072: cube 70  3073: tworoom 90  94/96 & tworoom 88/90/88 & 88/82 & 74/76
    \hline
    \multirow{3}{*}{Memory Retrieval Temperature}& 0.2 & 93.3 $\pm$ 0.94 & 90.0 $\pm$ 1.63 & 88.1 $\pm$ 5.85 & 74.7 $\pm$ 2.49 & 86.5\\ %94/94/92 & 92/90/88 & & 72/74/78
                                        & 0.5 & \textbf{96.7 $\pm$ 0.94} & \textbf{90.7 $\pm$ 1.89} & \textbf{90.7 $\pm$ 0.94} & \textbf{81.3 $\pm$ 1.89} & \textbf{89.9}\\ %cube 84/80/ avg 91.0
                                        & 1.0 & \underline{93.3 $\pm$ 2.49} & \underline{90.7 $\pm$ 0.94} & \underline{89.3 $\pm$ 2.49} & \underline{80.0 $\pm$ 1.15} & \underline{88.3}\\ %96/90/94 & 92/90/90 & 92/86/90& 78/80
    \bottomrule
  \end{tabular}
  \label{tab:mem}
\end{table*}

\begin{table*}[!t]
  \centering
  \small
  \setlength{\tabcolsep}{6pt}
  \renewcommand{\arraystretch}{1.12}
  \caption{Ablation on conditioning methods. We compare our default sequential design against the reverse and the separate conditioning alternatives. Best per column in \textbf{bold} and second-best \underline{underlined}.}
  \begin{tabular}{lccccc}
    \toprule
    \rowcolor{CornflowerBlue!10}
    {Method} & {Push-T (\%)}& {Two-Room (\%)} & {Reacher (\%)} & {OGBCube (\%)} & Average (\%) \\
    \hline
    Sequential (default) & \textbf{96.7 $\pm$ 0.94} & \textbf{90.7 $\pm$ 1.89} & \textbf{90.7 $\pm$ 0.94} & \textbf{81.3 $\pm$ 1.89} & \textbf{89.9} \\
    Reverse & 90.0 $\pm$ 1.63 & 87.3 $\pm$ 0.94 & \underline{86.7 $\pm$ 5.25} & \underline{69.3 $\pm$ 1.89} & \underline{83.3} \\
    Separate & \underline{90.7 $\pm$ 2.31}  & \underline{87.3 $\pm$ 3.06} & 84.7 $\pm$ 1.15 & 68.7 $\pm$ 2.31 & 82.9 \\
    \bottomrule
  \end{tabular}
  \vspace{-5pt}
  \label{tab:condition}
\end{table*}

\begin{figure}[!t]
\centering
\includegraphics[width=\linewidth]{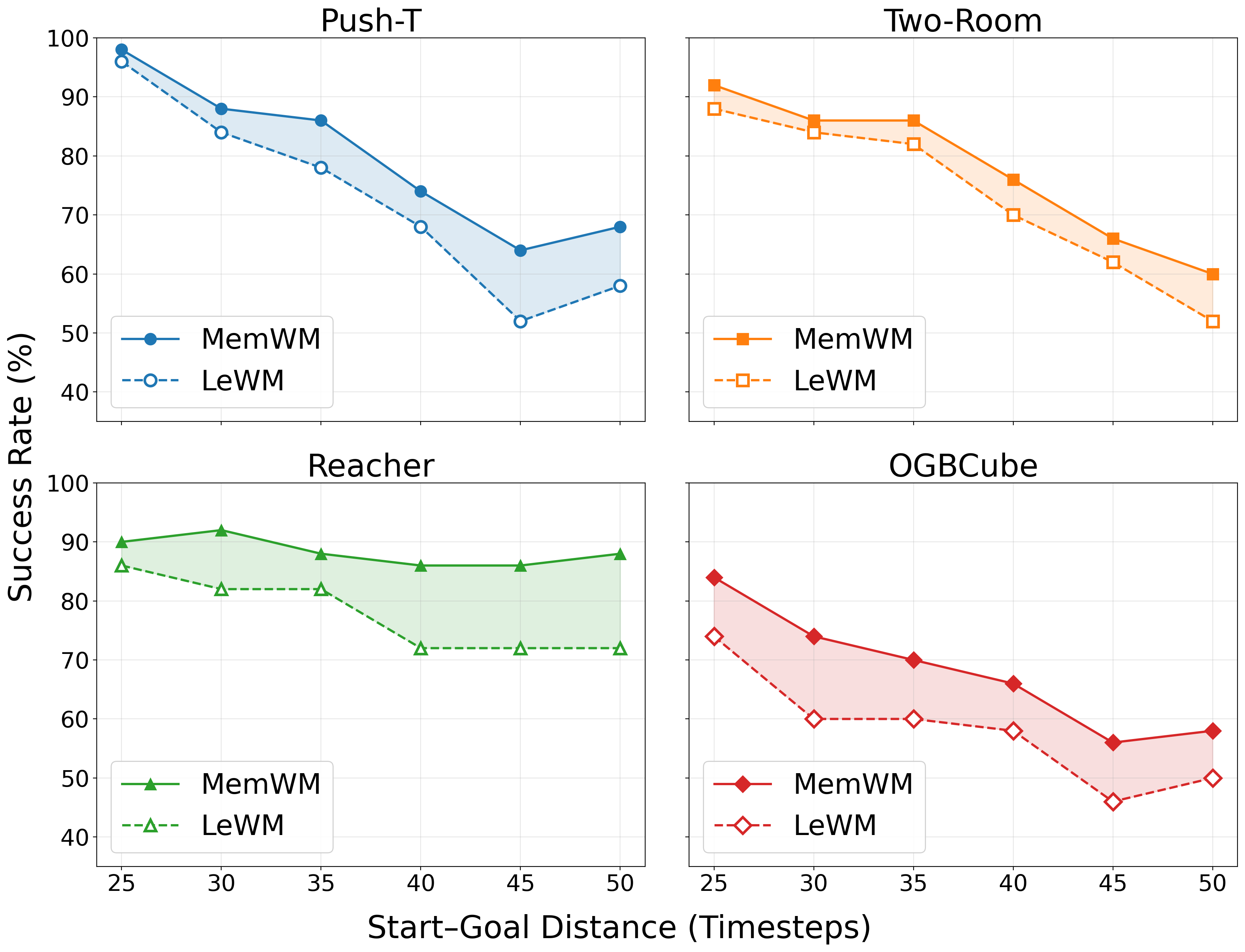}
\caption{Success rate as a function of start-goal timestep distance in the extended-horizon evaluation. As the horizon increases, the success rate degrades more rapidly for LeWM than for MemWM, indicating that associative memory conditioning mitigates error accumulation and enhances robustness over long horizons.}
%showing that MemWM degrades more gracefully over long horizons
\vspace{-10pt}
\label{fig:longhorizon}
\end{figure}

\paragraph{Discussion on conditioning methods}
\label{sec:condition}

To better understand the design choice of the conditioning method, we conduct ablation studies on how short-horizon action conditioning and long-horizon memory conditioning are combined within the predictor. We compare our default sequential design against two alternative strategies:
\begin{itemize}
    \item Reverse: We reverse the order of long-horizon memory conditioning and short-horizon action conditioning in AdaLN while maintaining the parameter sharing method.
    \item Separate: We employ two sequential AdaLNs with corresponding predictors. We then process short-horizon and long-horizon conditioning information through the two separate predictors, where the first predictor's output serves as input to the second. 
    %This is equivalent to the case of not using shared parameters.
    %\item Parallel: We leverage two separate AdaLN modules and predictors to process observations in parallel, conditioned on short-horizon actions and long-horizon memory contexts, respectively. The outputs of the two predictors are then summed to produce the next-step prediction.
\end{itemize}

%\paragraph{Success rate vs. test-time compute}

%We vary the planning budget by sweeping CEM samples and iterations. Figure \ref{} demonstrates the success rate in response to the increase of test-time compute, where the proposed method can achieve better performance at a similar computational cost comparing with other existing methods. Memory-enhanced latent world models achieves higher success while using up to .. less test-time compute than Hierarchical model planners.

\paragraph{Robustness to longer horizons}

To stress-test long-horizon planning, we evaluate an extended-horizon regime with start–goal pairs separated by up to 50 timesteps, double the 25 timesteps in the original evaluation. As shown in Figure \ref{fig:longhorizon}, LeWM exhibits a pronounced drop in success at longer horizons, indicating limited robustness to long-horizon generalization. MemWM degrades more gracefully, especially on Reacher, retaining 88.0\% success at 50 timesteps compared to 90\% at 25 timesteps, mitigating error accumulation over long horizons by conditioning on high-level memory information learned from historical data.

%goal-offset 40: pusht 62 reacher 78
%goal-offset 35:
%goal-offset 30:
%default goal-offset 25: pusht 96.7 tworoom 90.7 reacher 90.7 cube 81.3

%pusht: 25/90, 30/78, 35/78, 40/62, 45/55, 50/60  98 86 86 74 64 68
%tworoom: 25/88, 30/82, 35/72, 40/72, 45/60, 50/54  92 86 82 76 66 58
%reacher: 25/82,30/86,35/80,40/80,45/80,50/84  90 92 88 86 86 88
%cube: 25/81,30/72,35/68,40/64,45/50,50/54   84 74 70 66 56 58

%lewm: 96 87 86 74
%pusht: 96 86 78 72 52 58
%tworoom: 88 84 82 70 64 52
%reacher: 86 82 82 72 72 72
%cube: 74 60 60 58 52 56

\begin{comment}
\paragraph{Planning cost comparison}

We evaluate the method's computational cost in Table \ref{tab:compute} for planning in unseen scenarios, where performance on the Push-T dataset is used as an example to demonstrate the lightweight cost of the proposed associative memory approach in the latent world model architecture.

\begin{table}[h]
\centering
\caption{Planning time comparison using Cross-Entropy Method (CEM) with 50 samples per iteration and 10 optimization steps.}
\label{tab:compute}
\begin{tabular}{lcc}
\hline
\textbf{Metric} & MemWM & LeWM \\
\hline
Planning CEM (s) & 65.27 & 45.62\\ %59.14 & 39.02, 6.136 & 6.598 
GFLOPs & & \\
\hline
\end{tabular}
\end{table}
\end{comment}

\section{Conclusion}

We presented MemWM, an associative memory-enhanced latent world model for long-horizon planning. Our method facilitates learning latent dynamics via a content-addressable memory, while maintaining a lightweight architecture. We proposed two complementary memory losses to encourage memory diversity and balanced temporal utilization, effectively preventing memory learning collapse. Extensive experiments on PushT, Two-Room, Reacher, and OGBCube demonstrate that MemWM is competitive with the strongest DINO-WM method, using roughly 4$\times$ fewer trainable parameters and no pretrained vision backbone, outperforming latent world models of comparable size.

\paragraph{Limitations}

%Training becomes unstable and performance degrades in latent world models as the planning horizon increases. Our method alleviates this to a large extent, however it requires action-labeled observations for end-to-end training, which incurs a cost in collecting labels. 
%A better solution might be training a policy network that learns different actions within the latent space of the world model, eliminating the need to collect each action during planning. Learning such actions through model training could be challenging due to both the prediction of next state of environment and the action could be unstable, leading to collapse of model learning.
Our method requires action-labeled trajectories for training, which are costly to collect at scale. Future work could replace action labels with a learned inverse model that infers the action that best explains an observed latent transition. However, learning such actions through model training is challenging, as both next-state and action prediction may be unstable, potentially inducing collapse in model learning. A key open question is whether the inverse and forward models co-adapt to a degenerate latent space.

%9. For the visualization of memory slot usage, currently only the PushT and Reacher tasks demonstrate interpretability, with each memory slot exhibiting diversified activation patterns. For the OGBCube and TwoRoom tasks, further analysis is required to enhance the interpretability
%10.Memory slot usages are not well distributed across the different stages of tasks, is there a better approach to diversifying memory slots to further enhance model performance? Any better way to obtain embeddings evolve semantic information such as using the VLM-based guidance during training. 

\bibliographystyle{unsrt}
\bibliography{references}

\clearpage
\appendix
\section{Datasets}
\label{app:dataset}

We provide a detailed description of the four tasks used in our evaluation: PushT, TwoRoom, Reacher, and OGBCube. All tasks are implemented within the stable-worldmodel framework \cite{maes2026stable}, which provides a unified interface for environment interaction, data collection, and model evaluation. For each task, we describe the observation space, the action space, and the task objective.

\paragraph{PushT}
PushT is a robotic manipulation task adapted from the popular benchmark. The environment consists of a 2D top-down view of a tabletop with a T-shaped block. The agent controls a circular end-effector, and the goal is to push the T-shaped block from a random initial pose to a randomly sampled target pose. The task is considered successful when the block's position and orientation are within a specified threshold of the target. The observation is a $96 \times 96 \times 3$ RGB image rendered from a top-down camera view. No proprioceptive or state information is provided to the agent. The action space is a 2-dimensional continuous space representing the desired change in the end-effector's $(x, y)$ position.

\paragraph{TwoRoom}
TwoRoom is a 2D navigation task that requires an agent to navigate through two connected rooms to reach a specified goal location. The environment is a top-down view of a floor plan consisting of two rectangular rooms connected by a narrow doorway. The agent is a point-mass entity that must learn to move from a random starting position in the first room to a random goal position in the second room. The observation is a $64 \times 64 \times 3$ RGB image showing the entire floor plan, the agent's current position, and the goal position. The action space is a 2-dimensional continuous space representing the desired velocity of the agent in the $(x, y)$ plane.
    
\paragraph{Reacher}
Reacher is a continuous control task that involves a 2-link robotic arm moving its end-effector to randomly appearing target points in a 2D plane. The agent must learn to control the joint torques of the arm to reach the target as quickly and accurately as possible. The observation is a $64 \times 64 \times 3$ RGB image rendered from a fixed side-view camera. The target point is visually indicated as a red sphere. The action space is a 2-dimensional continuous space representing the torques of the two joints.

\paragraph{OGBCube}
OGBCube is a 3D manipulation task that requires an agent to interact with a cube in a simulated environment. The agent controls a robotic gripper and must learn to perform simple manipulation tasks such as pushing the cube to a target location or picking it up. The environment includes realistic physics and rendering. The observation is a $128 \times 128 \times 3$ RGB image rendered from a camera positioned above the workspace. The agent also receives proprioceptive information about the gripper's position and orientation. The action space is a 4-dimensional continuous space representing the desired change in the gripper's $(x, y, z)$ position and its yaw angle.

\section{Hyperparameters}
\label{app:hype}

We provide a detailed description of the hyperparameters employed in this work (Table \ref{tab:hyperparameters}). 

\begin{table}[h]
\setlength{\tabcolsep}{26pt}
\centering
\small
\begin{tabular}{lc}
\toprule
\textbf{Hyperparameter} & \textbf{Value} \\
\midrule
\multicolumn{2}{c}{\textbf{Training Configuration}} \\
\hline
Batch Size & 128 \\
Max Epochs & 100 \\
Train Split & 0.9 \\
Gradient Clipping & 1.0 \\
Image Size & 224 \\
Embedding Dimension & 192 \\
History Size & 3 \\
SIGReg Knots & 17 \\
SIGReg Projections & 1024 \\
SIGReg Loss Weight & 0.09 \\
\midrule
\multicolumn{2}{c}{\textbf{Optimizer}} \\
\hline
Optimizer Type & AdamW \\
Learning Rate & $5 \times 10^{-5}$ \\
Weight Decay & $1 \times 10^{-3}$ \\
\midrule
\multicolumn{2}{c}{\textbf{Memory Mechanism}} \\
\hline
Number of Slots & 10 \\
Memory Slot Dimension & 192 \\
Key Dimension & 192 \\
Temperature & 0.5 \\
Slot Initialization Scale & 0.02 \\
Key Initialization Scale & 0.02 \\
\midrule
\multicolumn{2}{c}{\textbf{Evaluation}} \\
\hline
Horizon & 5 \\
Receding Horizon & 5 \\
Frame Skip & 5 \\
Goal Offset Steps & 25 \\
Evaluation Budget & 50 \\
\bottomrule
\end{tabular}
\caption{The detailed settings for the memory bank component, the latent world model component, the training objectives, and the different baseline methods.}
\label{tab:hyperparameters}
\end{table}

\section{Additional results of planning trajectories}

\begin{figure}[h]
    \centering
    \includegraphics[width=\linewidth]{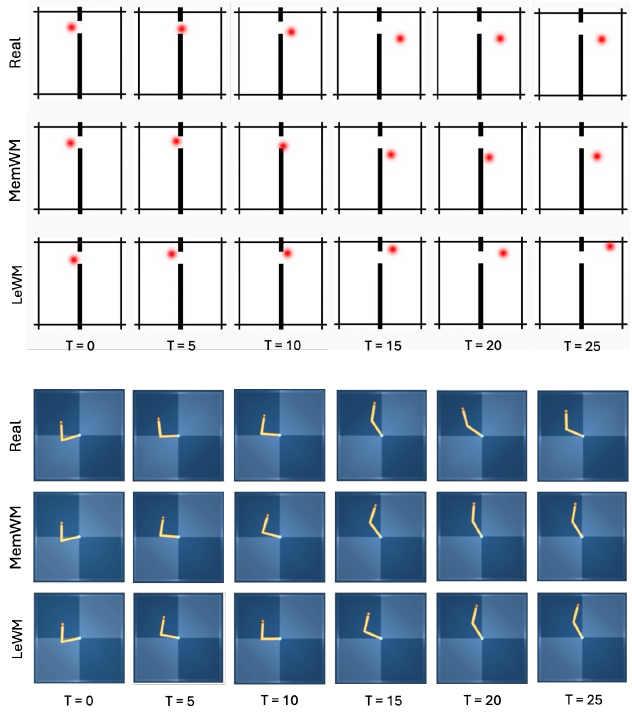}
    \caption{Additional visualization of planning trajectories based on MemWM.}
    \label{app:samples}
\end{figure}

We provide additional planning trajectory results for the TwoRoom and Reacher tasks in Figure \ref{app:samples}. Compared to LeWM, MemWM produces a more consistent trajectory rollout that is better aligned with the real dataset samples. In LeWM, the agent tends to jitter around uselessly due to the lack of high-level temporal information, using only the environment's low-level contact dynamics for conditioning. In contrast, MemWM leverages associative memory as additional conditioning information for the predictor, resulting in smoother transitions across timesteps.

\end{document}